\documentclass[manuscript,screen,nonacm]{acmart} 

\usepackage{subfigure}
\newtheoremstyle{case}{}{}{}{}{\bfseries}{.}{.5em}{\thmnote{#3}}
\theoremstyle{case}
\newtheorem*{case}{}

\AtBeginDocument{%
  \providecommand\BibTeX{{%
    \normalfont B\kern-0.5em{\scshape i\kern-0.25em b}\kern-0.8em\TeX}}}

\begin{document}

\title{Are There Exceptions to Goodhart's Law?}
\subtitle{On the Moral Justification of Fairness-Aware Machine Learning}

\author{Hilde Weerts}
\affiliation{
    \institution{Eindhoven University of Technology}
    \country{The Netherlands}
}
\email{h.j.p.weerts@tue.nl}

\author{Lamb\`{e}r Royakkers}
\affiliation{
    \institution{Eindhoven University of Technology}
    \country{The Netherlands}
}
\email{l.m.m.royakkers@tue.nl}

\author{Mykola Pechenizkiy}
\affiliation{
    \institution{Eindhoven University of Technology}
    \country{The Netherlands}
}
\email{m.pechenizkiy@tue.nl}

\renewcommand{\shortauthors}{Weerts et al.}
\begin{abstract}
Fairness-aware machine learning (fair-ml) techniques are algorithmic interventions designed to ensure that individuals who are affected by the predictions of a machine learning model are treated fairly. The problem is often posed as an optimization problem, where the objective is to achieve high predictive performance under a quantitative fairness constraint. However, any attempt to design a fair-ml algorithm must assume a world where Goodhart's law has an exception: when a fairness measure becomes an optimization constraint, it does \textit{not} cease to be a good measure. In this paper, we argue that fairness measures are particularly sensitive to Goodhart's law. Our main contributions are as follows. First, we present a framework for moral reasoning about the justification of fairness metrics. In contrast to existing work, our framework incorporates the belief that whether a distribution of outcomes is fair, depends not only on the cause of inequalities but also on what moral claims decision subjects have to receive a particular benefit or avoid a burden. We use the framework to distil moral and empirical assumptions under which particular fairness metrics correspond to a fair distribution of outcomes. Second, we explore the extent to which employing fairness metrics as a constraint in a fair-ml algorithm is morally justifiable, exemplified by the fair-ml algorithm introduced by \citet{hardt2016equality}. We illustrate that enforcing a fairness metric through a fair-ml algorithm often does not result in the fair distribution of outcomes that motivated its use and can even harm the individuals the intervention was intended to protect.
\end{abstract}

\maketitle

\section{Introduction}\label{sec1}
Fairness-aware machine learning (fair-ml) techniques are algorithmic interventions designed to ensure that individuals who are affected by the predictions of a machine learning model are treated fairly. In the computer science literature, learning a predictor that incorporates fairness considerations is typically formulated as an optimization problem, where the objective is to achieve high predictive performance under some quantitative fairness constraint. For example, a fair-ml technique can be applied to ensure that a machine learning model has similar accuracy across demographic groups. A corresponding fairness constraint could be a maximum difference in accuracy between groups.

There are various ways in which this optimization problem can be approached, each with different underlying assumptions and outcomes~\citep{kamiran2013techniques}. Pre-processing techniques aim to mitigate biases in the training data to mitigate downstream model unfairness \citep[][]{Kamiran2011, Zemel2013, feldman2015certifying}. Constrained learning approaches incorporate fairness constraints into the machine learning process, either by directly incorporating a constraint in the loss function \citep[][]{calders2009building, Kamiran2011, zafar2017fairness, zhang2018mitigating} or by learning an ensemble of predictors \citep[][]{agarwal2018reductions}. Post-processing algorithms adjust existing machine learning models to satisfy fairness constraints, either by adjusting the parameters of a trained model directly \citep{kamiran2010discrimination} or by post-processing the predictions of the model \citep{hardt2016equality}.

There is still little guidance on the suitability of specific fair-ml techniques across contexts. While a growing body of work considers the circumstances and moral frameworks under which the use of particular quantitative fairness metrics can be justified \citep{friedler2016possibility, Heidari2019, Card2020, leben2020normative, Reuben2020, Hertweck2021, hedden2021statistical}, the moral implications of the use of fair-ml techniques to satisfy fairness constraints remains largely unexplored. The distinction between fairness \textit{metrics} and \textit{constraints} is important. Approaching fairness as an optimization problem must assume a world in which Goodhart's law\footnote{ Goodhart's law, as popularized by \citet{strathern1997improving}, states that "\textit{when a measure becomes a target, it ceases to be a good measure}".} has an exception: when a fairness measure becomes an optimization constraint, it does \textit{not} cease to be a good measure. In this paper, we explore the moral space in which fairness metrics (do not) lose their value when used as a constraint. Our main contributions are as follows. 

To know whether a fair-ml algorithm achieves its intended goal, we first need to understand what it means for a classifier to be ``fair". To this end, we first present a framework for moral reasoning about the justification of fairness metrics. Previous work \citep{friedler2016possibility,Hertweck2021} has used egalitarian principles to argue that whether a distribution of outcomes is fair depends in part on the \textit{cause} of observed inequalities. We extend this work to illustrate that the fairness of a distribution of outcomes also depends on the \textit{utility} of an outcome for the decision subject and what moral claim they have to receive the related benefit or avoid the burden. We apply our framework to uncover a set of moral and empirical assumptions that could justify two popular fairness metrics under specific circumstances. 

Subsequently, we analyze the effectiveness of fair-ml under the identified assumptions. As a running example, we analyse the moral implications of the post-processing approach introduced by \citet{hardt2016equality}. Our arguments show that fairness metrics are typically simplistic operationalizations of more nuanced philosophical concepts of equality and discrimination. As a result, fair-ml algorithms do not always translate to the equal distribution of benefits and burdens that motivated their use.

To the best of our knowledge, we are the first to explicitly extend moral reasoning from fairness metrics to the fair-ml algorithms that optimize for them. For the avoidance of doubt, we do not aim to provide a complete set of precise and general statements regarding the applicability of fairness metrics and constraints. Instead, our work illustrates the type of reasoning that can be used to determine the suitability of fairness constraints. In part, our contribution can thus be viewed as a meta-ethical approach, relevant to a wider audience of computer scientists, philosophers, and social theorists interested in the impact of artificial intelligence on society. More generally, we hope our work opens up the discussion on the moral implications of fair-ml beyond fairness metrics.

The remainder of this paper is structured as follows. Section~\ref{sec:preliminaries} covers preliminaries, including three running examples that will be used throughout the paper, two fairness metrics, and a brief introduction to egalitarianism - the philosophical theory that underpins many of our arguments. In Section~\ref{sec:framework}, we describe a framework introduced by \citet{Hertweck2021} and propose an extension that considers the utility of predicted outcomes. In Section~\ref{sec:distilling}, we use our framework to identify circumstances where a particular fairness metric may (not) be appropriate. Subsequently, Section~\ref{sec:goodhartslaw} discusses the moral implications of using fair-ml to achieve a fairness metric, exemplified by the post-processing algorithm proposed by \citet{hardt2016equality}. We conclude with a discussion on the implications of our results (Section~\ref{sec:conclusions}).

\section{Preliminaries}
\label{sec:preliminaries}
In this section, we introduce notation and delineate required background information (Section~\ref{sec:examples}). First, we introduce three running examples that will be used throughout this paper. We then formally define two fairness metrics, demographic parity and equalized odds (Section~\ref{sec:metrics}). Finally, we provide a brief introduction to egalitarianism, the philosophical theory that provides the foundation of our framework (Section~\ref{sec:egalitarianism}). 

\subsection{Running Examples}
\label{sec:examples}
Our arguments will be illustrated using three running examples. Albeit stylized, these examples highlight key differences across scenarios that are relevant to our arguments.
\begin{quote}
\vspace*{-\baselineskip}
\begin{case}[Resume selection]
\label{case:resumeselection}
A machine learning model is used to select job applicants for interviews, based on their resumes. Due to resource constraints, only a limited number of interviews can take place. Based on the model's predictions, the top-ranking candidates are selected for an interview round. 
We define a candidate to belong to the \textit{positive} class when they have the qualities to be a high-performing employee. 
The model is trained to predict appraisal scores of current employees, based on employees' resumes at the time of application. 
We assume all candidates are sufficiently qualified such that they would benefit from getting the interview.
\end{case}
\end{quote}

\begin{quote}
\vspace*{-\baselineskip}
\begin{case}[Lending]
\label{case:lending}
The lending scenario considers a machine learning model that predicts whether an applicant will default on their loan within the first year, based on the characteristics of the requested loan and prospective borrower. A \textit{positive} prediction corresponds to the applicant receiving the loan and a \textit{negative} prediction to a rejection. The predictions will be used to automatically accept or reject loan applications. We further assume that there is no direct resource constraint, but that the lender wants to balance the false positives and false negatives to manage the overall risk. Finally, we assume that it is beneficial for the applicant to get a loan if the applicant can pay off the loan in time (\textit{true positive}), but harmful for an applicant who will default (\textit{false positive}).
\end{case}
\end{quote}

\begin{quote}
\vspace*{-\baselineskip}
\begin{case}[Disease detection]
\label{case:diseasedetection}
In the disease detection scenario, it is assumed that a machine learning model is developed to predict whether somebody is highly likely to get a deadly disease. We use \textit{positives} to refer to patients who have the disease and \textit{negatives} to patients who do not. The model is trained using historical patient data and is used to determine whether a patient will receive preventive treatment. However, the treatment has severe side effects, meaning that the patient's benefit and harm depend on the accuracy of the prediction.
\end{case}
\end{quote}

\subsection{Two Group Fairness Constraints}
\label{sec:metrics}
The majority of fair-ml algorithms, including the post-processing algorithm introduced by \citet{hardt2016equality}, optimize for \textit{group fairness}. Group fairness is a notion of algorithmic fairness that requires group statistics over predictions to be equal across (sub)groups, defined by one or more sensitive characteristics. Typical examples of sensitive characteristics are race and gender, but depending on the context other traits may be socially salient. In this work, we consider two fairness constraints for which the algorithm by \citet{hardt2016equality} can optimize: \textit{demographic parity} and \textit{equalized odds}.

We use the following notation. $Y$ denotes the target variable, which represents the outcome of interest that is to be predicted. $\widehat{Y}$ denotes the model's predictions. $X$ denotes the features on which the model's predictions are based. Finally, $A$ represents the (set of) sensitive features, which operationalize the sensitive characteristics of interest. For ease of presentation, the metrics will be illustrated for the binary classification scenario and a single sensitive feature.

\subsubsection{{Demographic parity}}
\label{sec:fairnessmetrics}
Demographic parity requires that the \textit{selection rate} is equal across sensitive groups. For a binary classifier, demographic parity is met if the following holds:
\begin{equation}
    \label{eq:demographicparity}
    P(\widehat{Y}=1 \mid A=a) = P(\widehat{Y}=1 \mid A=a'),
\end{equation}
for all $a, a' \in A$. For example, in \nameref{case:lending}, demographic parity is met if the proportion of applicants who receive a loan is equal across groups. 

Note that demographic parity does not depend on the target variable $Y$, only on the model's predictions. Consequently, when base rates differ between groups, i.e., $P(Y=1 \mid A=0) \neq P(Y=1 \mid A=1)$, demographic parity rules out a predictor with perfect accuracy.

\subsubsection{Equalized odds}
Equalized odds requires that the misclassification rates of the model are equal across all sensitive groups. In particular, it requires the \textit{false positive rate} and \textit{true positive rate} to be the same for each sensitive group. For a binary classifier, equalized odds is satisfied if the following holds:
\begin{equation}
    \label{eq:equalizedodds}
    P(\widehat{Y}=y \mid A=a, Y=y) = P(\widehat{Y}=y \mid A=a', Y=y),
\end{equation}
for all $a, a' \in A$ and  $y \in \{0,1\}$. In \nameref{case:diseasedetection}, the false positive rate constitutes the proportion of people who will \textit{not} get the disease (negatives) for which the model falsely predicts that they will get the disease (false positives). The true positive rate, on the other hand, amounts to the proportion of people who will get the disease (positives) that are correctly predicted to get the disease (true positives). 

As opposed to demographic parity, equalized odds conditions on the target variable $Y$. At first glance, disparities in false positive rates and true positive rates primarily suggest something about the predictive performance of the model. Indeed, if a model is generally more accurate for some groups compared to others, misclassification rates will be comparatively lower for those groups as well. However, equalized odds is also be affected by disparities in base rates: if the proportion of positives in a group is high, the distribution of false positives and false negatives will typically be different compared to a group with a relatively lower base rate. That is, even if the overall accuracy is similar, the distribution of different types of mistakes can differ, resulting in unequal odds.

\subsection{Egalitarianism}
\label{sec:egalitarianism}
Group fairness metrics require some form of equality across groups. Given this characteristic, several scholars have suggested that the philosophical perspective of egalitarianism may provide an ethical framework to understand and justify group fairness metrics \citep{binns2018fairness, Heidari2019, Hertweck2021}. In this section, we provide a brief introduction to some of the basic principles of egalitarianism.

Egalitarianism is a school of thought in political philosophy about \textit{distributive justice}: the just allocation of benefits and burdens.
Egalitarian theories are generally grounded in the idea that all people are equal and should be treated accordingly. 
An important concept within egalitarianism is \textit{equality of opportunity}: the idea that (1) social positions should be open to all applicants who possess the relevant attributes, and (2) all applicants must be assessed only on relevant attributes \citep{arneson2018four}. 
We can distinguish between formal and substantive interpretations of equality of opportunity.

A \textit{formal} interpretation of equality of opportunity requires all people to formally get the opportunity to apply for a position. Applicants are to be assessed on their merits, i.e., according to appropriate criteria. In particular, (direct) discrimination based on arbitrary factors, particularly sensitive characteristics such as race or gender, is prohibited. Note that formal equality of opportunity does not require applicants of all sensitive groups to have a non-zero probability to be selected. In particular, formal equality of opportunity allows the use of criteria that are (highly) related to sensitive characteristics, provided these criteria are deemed relevant for assessing merit. \textit{Substantive} theories go one step further and pose that everyone should also get a substantive opportunity to \textit{become} qualified. Two prominent substantive interpretations are \textit{Rawlsian equality of opportunity} and \textit{luck egalitarianism}. According to John Rawls' theory of justice \citep{rawls2020theory}, everyone with similar innate talent and ambition should have the same prospects for success, irrespective of their socio-economic background. Additionally, Rawls only considers social-economic inequalities acceptable if these inequalities benefit the most disadvantaged members of society. Luck egalitarianism \citep{dworkin1981equality}, on the other hand, poses that inequalities are only just if they are the result of informed, free choice, as opposed to brute luck \citep{binns2018fairness}. For example, taking a gamble with known risks is an informed choice, while being born into a rich household is the result of luck. Varieties of luck egalitarianism differ in how to distinguish between `luck' and `choice'.
As with any philosophical theory, egalitarianism has been the subject of criticism. Most objections revolve around the central egalitarian view that the presence or absence of inequality is what matters for justice. Alternative principles of distributive justice include, for example, \textit{maximin}, which requires the expected welfare of the worst-off group to be maximized. In particular, anti-egalitarianist philosophers often invoke the \textit{leveling down objection} \citep{parfit1995equality}, which points out that equality can be achieved through lowering the welfare of the better-off to the level of the worse-off, without making those worse-off any better-off in absolute terms.

Setting aside these objections for now, a primary question that remains is what decision subjects morally deserve, i.e., \textit{what} should be equal? When we consider fairness metrics through an egalitarian lens, we can see that each metric provides a different answer to this question. Enforcing demographic parity (Equation~\ref{eq:demographicparity}) implies that each group of individuals is, on average, equally deserving of $\widehat{Y}$, regardless of their ground-truth class $Y$. Equalized odds (Equation~\ref{eq:equalizedodds}), on the other hand, assumes that each group of individuals with the same ground-truth class $Y$ are, on average, equally deserving of $\widehat{Y}$. A direct consequence of these different valuations of predicted outcomes is that demographic parity and equalized odds are mathematically incompatible \citep{kleinberg2016inherent}: in many cases, it is impossible to achieve demographic parity and equalized odds simultaneously.\footnote{To be precise, demographic parity and equalized odds cannot hold at the same time if base rates differ across groups (i.e., $P(Y=1 \mid A=0) \neq P(Y=1 \mid A=1)$) and predictions $\hat{Y}$ are not independent of target variable $Y$.} The question of interest, thus, is under what conditions decision subjects have a moral claim to a particular distribution of outcomes.

\section{A Framework for the Justification of Fairness Metrics}
\label{sec:framework}
In this section, we introduce our framework that helps to reason about the moral and empirical assumptions under which the use of demographic parity and equalized odds may be justified from a moral perspective. 
We first introduce the framework proposed by \citet{Hertweck2021} (Section~\ref{sec:spaces}), which distinguishes between different causes of inequalities to justify the use of demographic parity. We then present an extension of this framework (Section~\ref{sec:extension}) which incorporates the belief that the fairness of the distribution of predicted outcomes depends on how the predicted outcome affects decision subjects.

\subsection{Framework by \citet{Hertweck2021}}
\label{sec:spaces}
The purpose of the framework introduced by \citet{Hertweck2021} is to identify circumstances under which the use of demographic parity (Equation~\ref{eq:demographicparity}) is justified. A key insight is that the \textit{cause} of observed inequalities matters, as this impacts decision subjects' moral claim to receive a benefit or avoid a burden. Notably, these causes cannot be deduced from the observed data alone and depend on the real-world context and data collection process.

\begin{figure}[ht]
\centering
    \subfigure[\label{fig:originalframework}The original framework by \citet{Hertweck2021}.]
        {\includegraphics[scale=0.5]{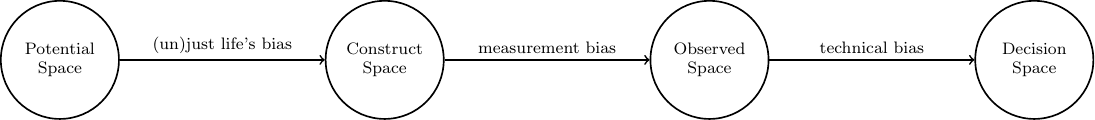}}
    \\
    \subfigure[\label{fig:adjustedframework}Our extension. The utility space is a function of the \textit{Decision Space} and, given the characteristics of the social context, may also depend on the \textit{Construct Space}.]
        {\includegraphics[scale=0.5]{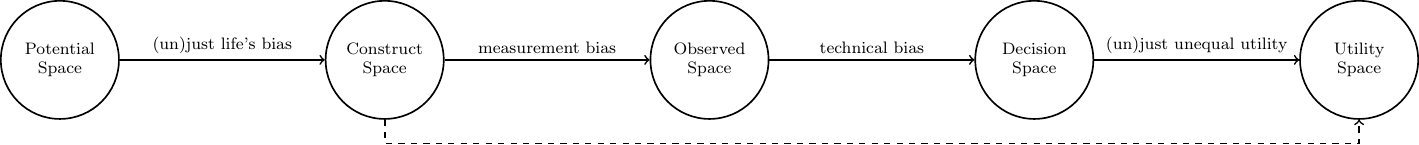}}
\caption{The relationship between the different spaces and distortion mechanisms. Each mechanism can cause inequality between groups going from one space to the next space.}
\label{fig:inequalityframework}
\end{figure}

The framework makes a distinction between four \textit{``spaces"}\footnote{In the original framework, the \textit{Potential Space}, \textit{Construct Space}, and \textit{Observed Space} are restricted to the characteristics we would like to base decisions on, whereas the DS considers the model's predictions. In this work, we explicitly include (predicted) \textit{outcomes} in the \textit{Potential Space}, \textit{Construct Space}, and \textit{Observed Space}. The reason for this is that a machine learning model is trained using features ($X$) \textit{as well as} a target variable \textit{Y}. As a result, measurement bias in the target variable can result in unfairness. For example, in \nameref{case:resumeselection}, predicting appraisal scores is a proxy for predicting employee quality. However, appraisal scores do not fully capture the substantive nature of employee quality, which threatens construct validity \citep{tambe2019artificial}. If historical appraisal scores are generated through biased evaluation practices, using the scores as a proxy for quality will constitute measurement bias. Borrowing the example of \citet{Reuben2020}, measurement bias could occur in \nameref{case:lending} if clerks are more lenient with repayment deadlines for men compared to women, resulting in more `late' payments for women. In \nameref{case:diseasedetection}, the disease may have been historically underdiagnosed in some demographic groups, causing measurement bias in the observed prevalence of the disease. We refer to \citep{Jacobs2021} for a more detailed overview of the relationship between measurement and fairness.}
that help to distinguish between different causes of observed inequality visualized in Figure~\ref{fig:originalframework}.

\begin{itemize}
    \item The \textit{Potential Space} represents an individual's innate potential to develop particular characteristics on which we would like to base decisions (we will refer to this as $\dot{X}$ for features and $\dot{Y}$ for the decision target). \textit{Example:} a person's potential to become knowledgeable about a particular job.
    \item The \textit{Construct Space} consists of the individual's materialized characteristics ($\tilde{X}$ for features and $\tilde{Y}$ for the decision target). \textit{Example:} a person's job knowledge.
    \item The \textit{Observed Space} consists of the measurements that are intended to measure the characteristics in the \textit{Construct Space} (i.e., the collected data set comprising of $X$ and $Y$). \textit{Example:} the number of years a person has worked in a similar job can be used as a proxy for a person's job knowledge.
    \item The \textit{Decision Space} represents the predictions of the machine learning model (i.e., $\widehat{Y}$), based on the measured characteristics in the \textit{Observed Space}\footnote{For simplicity, the \textit{Decision Space} is equated to the predictions of the machine learning model. We would like to emphasize that in reality, decisions will often be the outcome of a multi-step decision-making process that is part of a larger sociotechnical system including (multiple) human actors and other predictive models (see also \citep{Selbst2018}).}. \textit{Example:} a person's predicted appraisal scores.
\end{itemize}
Biases can distort transformation from one space to another space, introducing or exacerbating inequalities between groups.

\subsubsection{Technical Bias}
We refer to distortions between the \textit{Observed Space} and \textit{Decision Space} as \textit{technical bias}\footnote{\citet{friedler2016possibility} refer to distortions between the \textit{Observed Space} and \textit{Decision Space} as \textit{direct discrimination}. To avoid confusion with the concept of direct discrimination in European Union non-discrimination law, we take inspiration from \citet{friedman1996bias} and refer to this type of distortion as technical bias.}. The intuition is that if two groups are, on average, equal in the \textit{Observed Space}, inequalities in the \textit{Decision Space} are due to differences in treatment by the machine learning model. For example, in \nameref{case:lending}, the probability that an individual will default on a loan could be equal for women and men (i.e., the base rate in the \textit{Observed Space}, $P(Y=y)$, is equal across groups). If a model trained on this data produces relatively more positives for women compared to men (i.e., the base rates in the \textit{Decision Space}, $P(\widehat{Y}=y)$, are \textit{not} equal between groups), technical bias was introduced by the machine learning procedure.

\subsubsection{Measurement Bias}
Distortions between the \textit{Construct Space} and \textit{Observed Space} are a form of \textit{measurement bias}: a systematic error in measurement. For example, a person's number of years of experience is a proxy for their actual knowledge (i.e., a {measurement} in the \textit{Observed Space}) and could over- or underestimate actual knowledge (i.e., the {construct} the proxy is supposed to measure, which lies in the \textit{Construct Space}). If measurement bias is associated with sensitive group membership, this can introduce or exacerbate inequalities between groups in the \textit{Observed Space}. For example, in \nameref{case:resumeselection}, social bias in past performance evaluations may have made it generally more difficult for women to receive high appraisal scores compared to equally knowledgeable men, resulting in differences in observed average performance.

\subsubsection{Life's Bias}
Distortions between the \textit{Potential Space} and \textit{Construct Space} are due to what \citet{Hertweck2021} refers to as \textit{life's bias}: inequalities in circumstances, such as the income of a person's parents, that influence whether an individual's potential materializes. For example, in \nameref{case:resumeselection}, a lack of access to high-quality education could deprive a talented student of getting excellent grades. Borrowing an example from \citet{Reuben2020}, in \nameref{case:lending}, women might be ``more likely to be single parents with unpredictable outgoings due to their dependants". Finally, in \nameref{case:diseasedetection}, some demographic groups may be at higher risk for developing the disease due to confounding factors such as poorer nutrition.

\subsection{The Extension: The Utility Space}
\label{sec:extension}
The central idea of the framework by \citet{Hertweck2021} is that the \textit{causes} of inequalities are relevant for judging fairness. However, the authors also identify a set of counterexamples that show their proposed arguments to justify demographic parity do not apply generally. In particular, a problem with the proposed framework is that it disregards how predicted outcomes affect decision subjects. In this section, we propose an extension of the framework aimed at alleviating this problem: the addition of the \textit{Utility Space}.

\subsubsection{Why Causes are Insufficient}
As displayed in Figure~\ref{fig:originalframework}, the final space in the original framework is the \textit{Decision Space}, implying that the distribution of predictions (i.e., positives and negatives) is what matters for fairness. Judging fairness based on the distribution of predictions alone may be appropriate when predictions directly correspond to generally beneficial outcomes. For example, in the \nameref{case:resumeselection} scenario, by stipulation, all candidates can benefit from a positive prediction, irrespective of their true class. From the perspective of the decision subject, the distribution of positive predictions could therefore be a relevant `currency' for equality. 

However, as also pointed out by \citet{Hertweck2021}, causes alone are insufficient when a predicted outcome \textit{cannot} be considered universally beneficial. For example, in the \nameref{case:diseasedetection} scenario, clearly, a positive prediction is only beneficial if it is a \textit{true} positive - a false positive would unnecessarily expose the patient to severe side effects. Here, it seems impossible to judge the fairness of a particular distribution of predictions $\widehat{Y}$, without also taking into account the ground truth class $Y$.

The importance of considering the distribution of harm and benefit associated with predicted outcomes is well-established. For example, \citet{crawford2017trouble} calls for distinguishing explicitly between different types of fairness-related harm: \textit{allocation harm}, referring to a skewed distribution of resources and opportunities, and \textit{quality-of-service harm}, which refers to a skewed distribution of errors across groups. In the philosophical literature, \citet{castro2019wrong} and \citet{long2021fairness} have argued that whether a risk of misclassification is defensible depends on the benefits and burdens associated with being misclassified and what moral claim individuals have to avoid a burden or receive a benefit.
\subsubsection{The Utility Space}
In light of the moral relevance of the impact of predicted outcomes on decision subjects, we propose to extend the framework with a \textit{Utility Space}, visualized in Figure~\ref{fig:adjustedframework}. This space represents the \textit{utility} of a particular prediction \textit{for the individual who is affected by the prediction}\footnote{We would like to stress that the \textit{Utility Space} does not constitute a utilitarian approach to fairness, where the goal is to maximize the sum of individual utilities for all parties. Instead, it serves as a way to conceptualize \textit{what} should be equal between groups, within an egalitarian framework. Additionally, we would like to acknowledge that the concept of `utility' can be controversial, as quantifying utility is notorious for the inherently political nature of assigning a (numerical) value to an outcome \citep{hu2020welfare, Leben2020, Card2020}. We believe this issue is at least partially alleviated by our focus on \textit{individual} utility for the person under classification, as opposed to the more general utility of a particular outcome for society.}. How should `utility' be defined? Considering algorithmic decision-making, a general notion of utility considers the \textit{net benefit or harm} of a predicted outcome for the decision subject. As usual in matters concerning fairness and discrimination, a more specific definition is highly contextual. In some scenarios, such as our \nameref{case:resumeselection} example, predicted outcomes directly affect the distribution of resources or opportunities. Taking the perspective of a sufficiently qualified applicant, getting a job interview corresponds to a net benefit. In this scenario, disparities in the \textit{Decision Space} could thus be directly interpreted as disparities in the \textit{Utility Space}. In other scenarios, the net harm or benefit may depend primarily on the accuracy of a prediction. For example, in \nameref{case:diseasedetection}, the net benefit or harm of being classified as having the disease depends on whether that prediction is correct. Similarly, in \nameref{case:lending}, the net benefit or harm of a loan depends on one's ability to pay off the loan in time. In these cases, utility thus depends on both the \textit{Decision Space} and the \textit{Construct Space} (Figure~\ref{fig:adjustedframework}). As a result, group disparities in the \textit{Decision Space} may disappear when we move to the \textit{Utility Space}.

\subsubsection{Unequal Utility}
At a group level, distortions between the \textit{Decision Space} and \textit{Utility Space} may arise if a particular predicted outcome (e.g., a (false) positive or (false) negative) has, on average, a different utility across groups. Put differently, different individuals may have a different utility function. Similar to life's bias, this can be the result of inequalities in circumstances.
For instance, the net benefit of a job interview is arguably higher for those who are currently unemployed compared to those who are employed. If there exist disparities in unemployment rates across groups, the average utility will consequently differ as well. Being approved for a financial loan may be more beneficial for groups with, on average, fewer financial means. Similarly, in \nameref{case:diseasedetection}, a false positive will generally constitute negative utility, but the consequences to physical well-being may be even worse for groups who, on average, are in poorer physical condition. As a result, (un)just unequal utility can exacerbate inequalities between groups moving from the \textit{Decision Space} to the \textit{Utility Space}.

\section{Distilling Moral and Empirical Assumptions from the Framework}
\label{sec:distilling}
The \textit{Utility Space} allows us to explicitly take into consideration the impact of predicted outcomes on decision subjects. A crucial question remains unanswered: under what conditions do decision subjects have a moral claim to a particular \textit{distribution} of utility? The question of how we should value a distribution mirrors the `equality of what?' debate in egalitarianism \citep{sen1980equality,binns2018fairness,hu2020welfare}. Different theories of distributive justice offer different answers to the question of the right `currency' of equality. In this section, we build upon the arguments put forward by \citet{Hertweck2021} to show how the extended framework can assist in making explicit the empirical and moral assumptions that could justify a particular fairness metric. In particular, we expose how different types of distortions can violate particular conceptualizations of equality.

\subsection{Life's Bias}
\label{malifebias}
We start our discussion with the first distortion in the framework: life's bias. One of the primary contributions of \citet{Hertweck2021} is the recognition that conceptions of substantive equality reject inequalities in materialized characteristics caused by what the authors refer to as \textit{unjust} life's bias: unjust social structures that prevent a person from developing their innate potential. For example, a Rawlsian interpretation of equality could pose that whether a person gets a job interview should depend on their innate talent and ambition rather than their personal circumstances\footnote{Additionally, inequalities should "emerge through institutions arranged in away that delivers the greatest expectations of social primary goods to society's least advantaged members" \citep{Hertweck2021}.}, while a luck egalitarian may argue that differences in utility are only allowed insofar as they are the result of choice rather than luck. \citet{Hertweck2021} apply the luck egalitarian distinction between `luck' and `choice' to distinguish between just and unjust life bias. If a distortion between the \textit{Potential Space} and \textit{Construct Space} is due to personal choice, the authors consider this \textit{just} life bias. If the distortion is due to circumstances (i.e., luck) this is considered \textit{unjust} life's bias. For example, in \nameref{case:lending}, we might assume that one's ability to pay off a loan is solely due to informed choices they have made in their life and conclude that inequality in the \textit{Construct Space} is just.\footnote{The validity of this argument, of course, depends on the extent to which this is a realistic assumption} 

\paragraph{Demographic Parity}
Given the distinction between just and unjust life's bias, \citet{Hertweck2021} continue to argue that if we assume that the base rates in the \textit{Potential Space} are equal (i.e., $P(\dot{Y}=1 \mid A=a)=P(\dot{Y}=1 \mid A=a')$) and differences in the \textit{Construct Space} are due to \textit{unjust} life's bias, demographic parity is a justifiable measure of fairness (i.e., $P(\widehat{Y}=1 \mid A=a)=P(\widehat{Y}=1 \mid A=a')$). Considering the framework, \citet{Hertweck2021} thus argue that a substantive conception of distributive justice would restrict the desired distribution of predicted outcomes (primarily) to the distribution of outcomes suggested by the \textit{Potential Space}. For example, in \nameref{case:resumeselection}, we have assumed all candidates are sufficiently qualified such that positive prediction can be considered beneficial. The problem present in the original framework becomes clear when we consider cases in which utility depends on one's materialized characteristics. 

Take for example \nameref{case:diseasedetection}. A luck egalitarian may argue that disparities in physical well-being are only allowed insofar as these are the result of choice rather than luck. Imagine that every individual has the same innate potential to develop the disease. However, individuals in group $A$ are at a higher risk for developing the disease, compared to individuals in group $B$, due to a lack of access to good nutrition. We assume this to be a case of unjust life's bias: group $A$ is over-represented in areas that have limited access to supermarkets with affordable and nutritious food. Moreover, members of group $A$ more often have a low socio-economic status due to historical injustice and they do not have the means to move to a different area.\footnote{Note that the structure of our example closely resembles the counterexample provided by~\citep{Hertweck2021}.} The luck egalitarian argument suggests that the distribution of physical well-being \textit{ought} to depend on the \textit{Potential Space}. However, for a decision subject, the net benefit of disease classification depends on the accuracy of that classification. Were we to account for unjust life's bias by enforcing demographic parity, we would be required to diagnose fewer individuals in group $A$, depriving them of preventative treatment. Put differently, in an unequal \textit{status quo}, physical well-being \emph{is already affected by unjust life's bias}. Consequently, restricting the distribution of predicted outcomes to the distribution suggested by the \textit{Potential Space} would fail to take into account existing differences between groups - resulting in an unequal distribution of utility.

We find ourselves in a conundrum. On the one hand, substantive equality suggests the distribution of utility ought to be equal across groups to the extent that differences in materialized characteristics are caused by unjust bias. At the same time, the net harm and benefit of a decision cannot be seen in isolation from the position a person holds in an already unequal society. The crux of the problem is that machine learning models are typically but a single step in more complex decision-making sequences situated in a broader sociotechnical context. If an individual's socio-economic background has affected their prospects for success in hiring (e.g., due to lack of access to education), controlling for unjust life's bias only once they apply for a job will not meaningfully contribute to substantive equality. Similarly, an invitation to a job interview does not guarantee that substantive egalitarian principles are adhered to for the remainder of the selection process. A minimal conception of substantive equality ensures that the (predicted) outcome does not \textit{worsen} an individual's position in society, implying that an appropriate measure of fairness takes into consideration the utility of predicted outcomes for individuals. When a predicted outcome is universally beneficial or harmful and base rates in the \textit{Potential Space} are equal, an equal distribution of utility conditional on an individual's potential arrives exactly at the result of \citet{Hertweck2021}: demographic parity. When utility \textit{does} depend on materialized characteristics, demographic parity no longer results in an equal distribution of utility.

\paragraph{Equalized Odds}
As equalized odds explicitly conditions on ground-truth variable $Y$, one may wonder whether this fairness measure would correspond to an equal distribution of utility when unjust life's bias has occurred and utility depends on materialized characteristics. Universally, this is not the case. For example, consider a scenario in which a utility function assigns the same utility to a false positive as to a false negative.\footnote{We note that in practice, false positives and false negatives are typically valued differently.} A classifier which is equally accurate for members of two groups, but misclassifies members of group $A$ strictly as false positives and members of group $B$ strictly as false negatives distributes utility equally, but violates equalized odds.

Importantly, under unjust life's bias, the prevalence of the positive class differs across groups, which usually affects the distribution of false positives and false negatives. For example, if the proportion of positives in group $A$ is higher than the proportion of positives in group $B$, a well-calibrated and reasonably accurate (but imperfect) classifier will misclassify instances in group $A$ relatively more often as a false positive, while instances in group $B$ will be misclassified relatively more often as a false negative. Put differently, when an error occurs, the \textit{type} of error that occurs is, on average, different for members of group $A$ compared to members of group $B$. This is a direct implication of the so-called impossibility result: if base rates differ between sensitive groups, a well-calibrated classifier cannot satisfy equalized odds \citep{kleinberg2016inherent,chouldechova2017fair}. 

Addressing unequal odds that are the result of disparities in base rates implies a utility function that values different types of mistakes differently across groups. For example, consider \nameref{case:lending}. If unjust life's bias has caused the default rate of women to be higher than that of men, women will on average be more likely to be misclassified to default. To equal the odds, we could thus decide to apply a more lenient lending policy for women compared to men. Such a policy implies we assign comparatively higher costs to false positives for women than for men. Such a policy could be justified if we take a substantial perspective on equality. In particular, individuals who belong to a group that is already disproportionately affected by historical injustice could have a reason to reject a classifier which reproduces an unfair status quo.\footnote{\citet{castro2019wrong} and \citet{long2021fairness} make similar arguments to explain why failing to satisfy equalized odds in pretrial risk assessment is morally wrong.} In this scenario, equalized odds could be justified as a measure of fairness.
 
\subsection{Measurement Bias}
\label{mameasurementbias}

We now consider distortions between the \textit{Construct Space} and the \textit{Observed Space} caused by measurement bias. \citet{Hertweck2021} argue that it is unfair to make decisions based on incorrect measurements: individuals cannot be held morally responsible for incorrect beliefs decision-makers have about them, so they should not have to bear harmful consequences of incorrect measurements. In the remainder of this section, we further substantiate this argument.

We believe the essence of the argument to be a formal conception of equality, closely following the Aristotelian dictum that "likes should be treated alike". For instance, incorrect measurements of an applicant's qualities could deprive them of the opportunity to participate in a job interview, while an equally qualified but correctly assessed applicant would be able to participate in the interview round. Vice versa, a less qualified applicant may be incorrectly classified as qualified, while an equally qualified but correctly assessed applicant is classified as unqualified. Similarly, biases in historical diagnoses could lead to disparities in misdiagnosis across sensitive groups. As a result, individuals who belong to a disadvantaged sensitive group would be disproportionately exposed to harmful effects of misdiagnosis compared to individuals with otherwise similar characteristics who belong to a different sensitive group.

Importantly, formal conceptions of equality assume that the net benefit or harm individuals experience ought to depend on one's materialized merit. That is, the "likeness" of individuals is characterized by the distribution of their materialized qualities in the \textit{Construct Space}. Under this worldview, individuals have a moral claim to equal outcomes insofar as that outcome is justified by their materialized characteristics. Considering the framework, formal equality thus can be said to restrict the distribution of predicted outcomes to the distribution suggested by the \textit{Construct Space}.

Substantive theories of justice would generally also reject inequalities caused by measurement bias, unless unequal treatment is applied to redress unequal circumstances. Under the assumption that any differences between groups arise from just life's bias, disaggregated measures of predictive performance may seem a reasonable fairness metric if we wish to achieve equal utility. However, under measurement bias, predictive performance metrics can be ruled out as a condition for fairness: predictive performance is computed based on measurements in the \textit{Observed Space}, which by stipulation does not accurately capture the \textit{Construct Space}. To measure equality under measurement bias, we must therefore make explicit assumptions regarding the true distribution of materialized characteristics in the \textit{Construct Space}. One of many possible assumptions is that group-specific base rates in the \textit{Construct Space} (i.e., $P(\tilde{Y}=1)$) are equal across groups. Under these specific circumstances, we arrive at the same conclusion as \citet{Hertweck2021}: a classifier that distributes utility equally across groups will exhibit demographic parity.

\subsection{Technical Bias}
\label{matechnicalbias}
Notions of formal equality could also provide a moral argument for decision subjects to object to a classifier which suffers from technical bias. In this case, distortions between the \textit{Observed Space} and \textit{Decision Space} are the result of differences in treatment by the machine learning model. For example, the model in \nameref{case:diseasedetection} may be less accurate for some groups compared to others, introducing disparities of physical well-being (\textit{Utility Space}) that were not present in the distribution of the `ground-truth' target variable (\textit{Observed Space}). Note that the classifier does not need to make (implicit or explicit) use of a sensitive feature for technical bias to occur. The provided features could simply be less informative for some groups. For example, a particular disease may present differently across biological sexes. As a result, a set of health indicators used in predictive modelling may be relevant factors for diagnosis in women, but much less useful to accurately diagnose men.

In contrast to measurement bias, technical bias does not affect the distribution of outcomes in the \textit{Observed Space}. Assuming the absence of measurement bias, the \textit{Observed Space} closely approximates the \textit{Construct Space}. As such, equalized odds is not ruled out as a fairness metric. Indeed, if base rates are equal between groups but the model is less accurate for some groups compared to others, misclassification rates will be higher, resulting in an unequal distribution of utility that cannot be explained by the person's materialized characteristics in the \textit{Construct Space}.

For example, consider the two group-specific ROC curves in Figure~\ref{fig:intersect}. Apart from trivial endpoints, the curves do not intersect: the ROC curve for group $A$ lies strictly above the ROC curve of group $B$. As a result, the diagnostic ability of the model is worse for group $B$ compared to group $A$. Assuming the prevalence of the positive class is equal across groups (see Section~\ref{malifebias}) at any decision threshold $t$, we will achieve a more favourable false positive rate and/or true positive rate for group $A$ than for group $B$. In this case, a violation of equalized odds thus violates a notion of formal equality.

\subsection{Unequal Utility}
\label{mautility}
We now arrive at the last distortion: unequal utility. Similar to unjust life's bias, a luck egalitarian conception of equality could distinguish between unequal utility due to informed choice (\textit{just} unequal utility) and unequal utility for which individuals should not be held accountable (\textit{unjust} unequal utility). For example, the consequences of a false negative in \nameref{case:lending} are worse for an individual who generally has a harder time getting a loan due to the unconscious bias of lenders, a factor related to luck rather than choice. On the other hand, if an individual purposefully withholds payment, unequal utility could be considered just. In the absence of unjust life's bias, measurement bias, and technical bias, accounting for unjust unequal utility must correspond to differential treatment. As we have seen in Section~\ref{malifebias}, a substantive conception of equality could provide a moral argument to value outcomes differently across groups. In this case, the relevant measure of fairness highly depends on the supplied utility function and cannot be connected to demographic parity or equalized odds in the general case.

\section{When a Fairness Measure Becomes a Constraint, Does it Cease to Be a Good Measure?}
\label{sec:goodhartslaw}
Having identified several empirical and moral assumptions under which demographic parity or equalized odds could be used as a measure of fairness, we now turn to the question of the extent to which employing these metrics as a (soft) constraint in a fair-ml algorithm is justifiable. To make our arguments concrete, we take as an example the post-processing algorithm proposed by \citet{hardt2016equality}. We first explain the post-processing algorithm and identify potential objections against deploying the technique. Subsequently, we explore the suitability of demographic parity and equalized odds as optimization constraints under the moral and empirical assumptions identified in the previous section.

\subsection{Group-Specific Randomized Decision Thresholds}
\label{sec:algorithm}
The goal of the post-processing algorithm introduced by \citet{hardt2016equality} is to identify a decision threshold%
\footnote{Many machine learning classification algorithms do not directly output a class, but a probability or score $R$, which indicates the confidence of the model that an instance belongs to a certain class. The decision threshold $t$ is the cut-off value of the model's predicted values at which you classify an instance as belonging to that class, i.e., if $R > t$, $\widehat{Y}=1$.} 
($t$) of the model's predicted scores ($R$) such that the post-processed predictions ($\widetilde{Y}$) adhere to a given fairness constraint while retaining as much predictive performance as possible. By moving the decision threshold up or down, we can tweak the trade-off between (false) positives and (false) negatives. The post-processing technique uses two specific strategies to satisfy the fairness constraints: group-specific thresholds and randomized thresholds.

\subsubsection{Group-Specific Decision Thresholds}
The trade-off between false positives and false negatives is directly related to the choice of decision threshold of a risk score function. A low decision threshold implies that many instances are classified as positive, resulting in more false positives compared to a more conservative, higher decision threshold. Literature on cost-sensitive classification tells us that, given a reasonable\footnote{The reasonableness conditions merely state that misclassification is more costly than a correct classification: $c(FP) > c(TN)$ and $c(FN) > c(TP)$.} cost function $c(A)$ of different types of outcomes and a calibrated risk score function $r(x)$, the optimal decision threshold $t^{*}$ is defined as \citep{elkan2011foundations}:
\begin{equation}
\label{eq:costsensitivethreshold}
    t^{*} = \frac{c(FP)-c(TN)}{c(FP)-c(TN) + c(FN)-c(TP)}
\end{equation}
In other words, provided that a risk score function is calibrated, a decision threshold represents a specific attitude regarding misclassification costs \citep{birch2022clinical}. Note that this is true, irrespective of whether the decision threshold is explicit (in the form of a classification model or decision-making policy with a predetermined decision threshold) or implicit (in the form of interpretations of the risk score by a decision-maker). Decision threshold selection is, therefore, an inherently context-dependent and normative decision.

If base rates differ across groups, it will often not be possible to identify one unique threshold $t$ such that a fairness constraint holds across all groups. In that case, \citet{hardt2016equality} propose to choose separate thresholds $t_a$ for sensitive groups $a \in A$. For example, to satisfy demographic parity (Equation~\ref{eq:demographicparity}), we could decrease the decision threshold for a group with a low selection rate, such that more instances are classified as positive. 

A common objection against group-specific thresholds is that it implies that each group is held to a different standard: $\widehat{Y}$ has a different meaning depending on sensitive group membership \citep{green2021escaping}\footnote{Note that applying a single decision threshold is not sufficient to ensure that the machine learning model treats members of different sensitive groups in the same way. For example, a sensitive feature may be used explicitly in the prediction-generating mechanism of the machine learning model. To simplify the current discussion, we assume that sensitive features are not used directly as model input, but the model can still (indirectly) rely on the sensitive feature through to the relationship (e.g., correlation) between the sensitive feature and other input features.}. Consequently, group-specific thresholds can be interpreted as a form of technical bias: it introduces a group-level distortion between the \textit{Observed Space} ($Y$) and the \textit{Decision Space} ($\widehat{Y}$). Group-specific thresholds thus seem to violate formal equality of opportunity: we base the decision on sensitive attributes which ought to be irrelevant to the prediction problem.

\subsubsection{Randomized Decision Thresholds}
\begin{figure}[ht]
    \centering
    \includegraphics{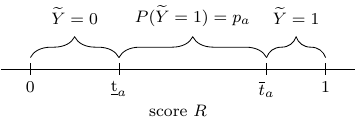}
    \caption{The randomized decision threshold for a probabilistic classifier. The randomized threshold $T_{a}$ is equal to $\underbar{t}_{a}$ with probability $p_{a}$ and equal to $\overline{t}_{a}$ with probability $1 - p_{a}$.}
    \label{fig:randomizedthreshold}
\end{figure}

\begin{figure}[ht]
    \subfigure[\label{fig:randomizedroccurve}Randomization of the decision threshold between values $\underbar{t}_{a}$ and $\overline{t}_{a}$ allows us to achieve any ($fpr$, $tpr$) combination on the line segment between the corresponding points on the ROC curve. The exact coordinate is determined by $p_a$, where higher values of $p_a$ are closer to $\overline{t}_{a}$ and vice versa.]
        {\includegraphics[width=0.3\textwidth]{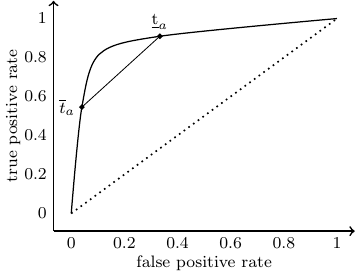}}
    \hfill
    \subfigure[\label{fig:intersect}The group-specific ROC curves of group $A$ (solid) and group $B$ (dashed) do not intersect, except at trivial end points. We can achieve equalized odds by randomizing the decision threshold of group $A$ such that it coincides with the ROC curve of group $B$.]
        {\includegraphics[width=0.3\textwidth]{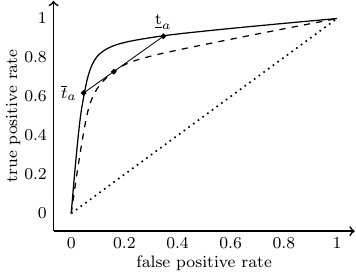}}
    \hfill
    \subfigure[\label{fig:suboptimal}The ROC curves of group $A$ (solid) and group $B$ (dashed) only intersect at a point with a high false positive rate (square). If the cost of false positives is high, this point may not be optimal. Randomization of the thresholds of group $A$ allows for a different trade-off which still satisfies equalized odds.]{
        \includegraphics[width=0.3\textwidth]{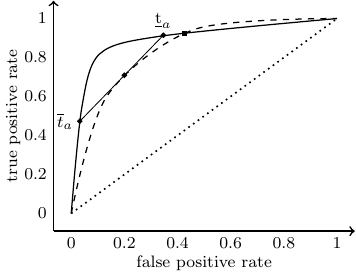}}
\caption{ROC curves of randomized decision thresholds.}
\label{fig:three}
\end{figure}

Group-specific thresholds are not always sufficient to achieve equalized odds. Recall that the main requirement of equalized odds is to ensure that the false positive rate and the true positive rate are equal across groups (Equation~\ref{eq:equalizedodds}). The trade-off between false positives and false negatives is often analyzed using a Receiver Operating Characteristic (ROC) curve, which sets out a classifier's false positive rate (\textit{fpr}) against its true positive rate (\textit{tpr}) over varying decision thresholds. Considering equalized odds, group-specific thresholds still limit us to the (\textit{fpr}, \textit{tpr}) combinations that lie on the intersection of group-specific ROC curves. In some cases, the group-specific ROC curves may not intersect (Figure~\ref{fig:intersect}) or represent a poor trade-off between false positives and false negatives (Figure~\ref{fig:suboptimal}). To further increase the solution space, \citet{hardt2016equality} allow the decision thresholds to be {randomized}. That is, the decision threshold $T_a$ is a randomized mixture of two decision thresholds $\underbar{t}_{a}$ and $\overline{t}_{a}$ (Figure~\ref{fig:randomizedthreshold}). 

Randomization allows us to achieve \textit{any} combination of (\textit{fpr}, \textit{tpr}) that lies within the convex hull of the ROC curve (Figure~\ref{fig:randomizedroccurve}). In cases where group-specific ROC curves do not intersect apart from trivial end points, the predictive performance of the model for the best-off group is artificially lowered through randomization until the performance is equal to that of the worst-off group (Figure~\ref{fig:intersect}). In cases where the ROC curves do intersect, but at a sub-optimal point, which group is affected depends on the specific trade-off that is considered (Figure~\ref{fig:suboptimal}).

While `flipping a coin' is a tried and tested approach for choosing between two competing alternatives or settling disputes. we may wonder whether randomization is justifiable as a means to achieve fairness.
A major objection against randomization in decision-making revolves around the belief that a good decision must be justified by explicit reasons \citep{keren2010decisions}. We can interpret this through Aristotle's concept of justice as consistency: "a decision-maker should be able to produce a single, predictable, and correct judgment in each case" \citep{Reuben2020}. As the gravity of the decision increases, such as in a case of life and death, a coin toss becomes less acceptable \citep{baron1997protected, beattie1994psychological}. In machine learning, randomization of predictions implies that two very similar or even identical instances can receive a different prediction, violating consistency. A related objection is that randomization makes it difficult to hold decision-makers accountable for the outcome \citep{keren2010decisions}. These objections bring into question the proportionality of randomization for achieving a particular distribution of burdens and benefits.\\

\noindent In Section~\ref{sec:distilling}, we have identified various moral and empirical assumptions under which demographic parity and equalized odds represent a justifiable distribution of burdens and benefits. In the remainder of this Section, we will consider to what extent fairness-aware machine learning and (randomized) group-specific decision thresholds in particular are an effective and justifiable way to achieve fairness.

\subsection{Unjust Life's Bias}
\label{sec:dplb}
In Section~\ref{malifebias}, we have distilled a set of assumptions under which demographic parity could be a relevant measure of fairness: if we assume that (i) base rates are equal in the \textit{Potential Space}, but unequal in the \textit{Construct Space} due to unjust life’s bias, and (ii) predicted outcomes are universally beneficial, a notion of substantive equality could require a classifier to satisfy demographic parity. Additionally, we argued that if predicted outcomes are not universally beneficial, equalized odds could be a relevant measure of fairness.

Group-specific decision thresholds are particularly well-suited to tweak the distribution of (false) positives and (false) negatives. If predicted scores are an accurate indication of an individual's materialized characteristics, the fair-ml algorithm therefore seems a suitable candidate to tackle unjust life's bias: we can simply lower or increase the decision threshold such that the desired distribution of utility is achieved.

The effectiveness of group-specific thresholds depends on the extent to which we understand how life's bias has materialized. For example, in \nameref{case:resumeselection}, disparities in opportunities regarding education may result in \textit{systematically} lower scores for a disadvantaged group, which could justify the use of a different threshold proportionate to the experienced bias. Similarly, in \nameref{case:lending}, we may use equation~\ref{eq:costsensitivethreshold} to carefully choose a set of group-specific decision thresholds which take into consideration assumed group-specific misclassification costs. In practice, identifying an appropriate correction can be challenging. For example, taking a luck egalitarian interpretation of unjust life's bias, an appropriate correction for life's bias must understand exactly how factors corresponding to either `choice' or `luck' affect materialized characteristics.

Importantly, like most fair-ml algorithms, the algorithm proposed by \citet{hardt2016equality} optimizes for a predictive performance metric (e.g., overall accuracy), under an equality constraint (demographic parity or equalized odds). While this may result in an equal distribution of (false) positives and (false) negatives, the approach is insensitive to any further assumptions regarding the desired distribution of burdens and benefits. As such, automatic selection of group-specific thresholds is not generally guaranteed to produce an equal distribution of utility.

\subsection{Measurement Bias}
\label{sec:dpmb}
In Section~\ref{mameasurementbias}, we have argued that under the assumption that base rates are unequal between groups due to measurement bias, a notion of formal equality could justify demographic parity as a measure of fairness. However, our finding does not imply that \textit{any} classifier which satisfies demographic parity corresponds to an equal distribution of benefits and burdens across groups. 

Consider for example \nameref{case:diseasedetection}. Imagine we are able to design a classifier to predict a positive if and only if receiving preventative treatment will be beneficial to the patient, given the available evidence. Suppose the classifier is the \textit{best} way to diagnose the disease, irrespective of group membership. However, individuals in group $A$ are more often asymptomatic. As a result, the disease is generally harder to detect with known methods for individuals in group $A$, causing the classifier to underestimate the prevalence of the disease in group $A$ relative to group $B$. Enforcing demographic parity in this scenario requires us to diagnose more individuals in group $A$, even though by stipulation there is no evidence that allows us to decide \textit{which} individuals will benefit from the preventative treatment. In other words, although a similar proportion of individuals would benefit from preventative treatment, enforcing similar treatment rates is not guaranteed to achieve an equal distribution of utility.

Now, consider \nameref{case:resumeselection}. Suppose that the appraisal scores are \textit{systematically lower} for a disadvantaged group $A$: all members of group $A$ have a similar probability of being subjected to unconscious bias and their appraisal scores are lowered by roughly the same amount. Under these circumstances, it is harder for members of group $A$ to get a certain score compared to equally qualified members of an advantaged group $B$. In this scenario, careful lowering of the decision threshold for members in group $A$, proportionate to the bias in appraisal scores, directly addresses measurement bias. Note that under this type of measurement bias, using a \textit{single} threshold in the \textit{Observed Space} implies the use of a \textit{different} threshold for each group in the \textit{Construct Space}. In this case, therefore, group-specific decision thresholds are best understood as a means to counteract existing violations of formal equality. As before, we find that the algorithm does not generally guarantee an equal distribution of utility and its effectiveness of highly depends on the way in which measurement bias has materialized.

\subsection{Technical Bias}
\label{sec:eo}
As explained in Section~\ref{matechnicalbias}, a notion of formal equality could justify equalized odds as a fairness metric if differences in misclassification rates are caused primarily by differences in predictive performance due to technical bias.

Machine learning approaches designed to increase predictive performance could be used to increase accuracy if deployed in a group-sensitive way sensitive groups. In particular, multi-objective optimization approaches could be deployed to ensure a model is sufficiently accurate across all groups \citep{weerts2024can}. In some cases, it may be impossible to increase predictive performance for the disadvantaged group. For example, the available data may not be sufficiently predictive.

Importantly, group-specific decision thresholds are unable to improve the predictive performance of a group: while it is possible to achieve any combination of false positive and true positive rates on the group-specific ROC curve, the choice of threshold cannot increase the \textit{area} under the ROC curves. Under predictive performance disparities, randomized group-specific thresholds can therefore only achieve equal misclassification rates by artificially decreasing performance for all groups until it is equal to that of the worst-off group. In this scenario, the approach invokes one of the main objections against egalitarianism: the levelling down objection.

In this scenario, randomized thresholds are unlikely to improve the absolute utility of the disadvantaged group -- unless utility depends on the predicted outcomes of individuals in other groups. For example, consider \nameref{case:resumeselection}. The resource constraint implies that a false negative in an advantaged group provides an opportunity for a member of the disadvantaged group. If a member of a disadvantaged group belongs to a historically marginalized group, misclassification may be considered more harmful compared to the misclassification of a member of an advantaged group (i.e., unequal utility). This difference in the costs of misclassification can provide decision subjects with a moral claim to avoid misclassification. For example, in \nameref{case:resumeselection}, one may argue that a member of group $A$ would be willing to forfeit a (true) positive prediction to compensate for the historical injustice of individuals in group $B$. In \nameref{case:diseasedetection}, we cannot make such an argument: unless we aim to achieve equality for the sake of equality, decreasing the predictive performance for group $A$ does not benefit individuals in group $B$.

This still leaves the question of the proportionality of randomization to achieve a particular distribution of burdens and benefits. As explained in Section~\ref{sec:algorithm}, randomization violates the principle of consistency of decision-making. However, some outcomes are inherently difficult to predict. For example, the construct of `merit' in the context of employment is complex and has an inherently stochastic nature \citep{tambe2019artificial}. Could the inherent uncertainty of the outcome that is predicted justify randomized decision-making?

\citet{hardt2016equality} implicitly incorporate this reasoning by applying randomization only if the predicted score is uncertain, i.e., if $\underbar{t}_{a} < R \leq \overline{t}_a$ (Figure~\ref{fig:randomizedthreshold}). However, we can identify at least two problems. First, an underlying assumption is that the predicted score of the classifier provides an estimate of whether an individual is a `border case'. While predicted scores will accurately reflect the inherent uncertainty of an outcome variable in some cases, the decision boundary of the classifier can also be very different from the `ground-truth' decision boundary. For example, a model optimized for classification accuracy can be poorly calibrated, meaning that predicted scores do not coincide with the probability of belonging to the positive class. Furthermore, the decision boundary of a model can be erratic due to e.g., overfitting (the model memorizes the training data and fails to generalize to unseen data) or small sample sizes (a particular subgroup may not be adequately represented in the data set, resulting in poor estimates for that group). Predicted scores, thus, may not accurately reflect real-world uncertainty. Second, the group-specific procedure does not put any restrictions on the width of the interval $[\underbar{t}_{a}, \overline{t}_{a}]$. At the extreme, we may find that any person in a group subject to randomization receives a random prediction, disregarding their personal characteristics and violating the principle of consistency. In other words, while randomization can be a useful trick to achieve a particular distribution of misclassification rates at a group level, it may not be proportionate from the perspective of the individual.

\section{Conclusions}
\label{sec:conclusions}
Any attempt to design a fair-ml algorithm which incorporates a fairness constraint must assume a world where Goodhart's law has an exception: when a fairness measure becomes an optimization constraint, it does \textit{not} cease to be a good measure. In essence, Goodhart's law calls attention to the inherent complexity of the world that makes it difficult to anticipate all effects of an intervention. Like any mathematical abstraction, fairness metrics are simplified formalisations of complex philosophical notions of fairness, discrimination, and equality. In this work, we have shown that this makes fairness metrics are particularly sensitive to Goodhart's law.

Our work presents two main contributions. First, we have presented a framework that can help identify moral and empirical assumptions under which the use of fairness metrics is justifiable. In contrast to previous work, our framework reflects that whether a distribution of outcomes is fair, depends in part on how an outcome affects decision subjects and what moral claims they have to receive the related benefit or avoid the burden. We have applied the framework to identify particular moral and empirical assumptions under which a fair classifier will satisfy two popular fairness metrics. Second, considering the identified assumptions, we have shown that simply \textit{enforcing} a fairness metric by means of a fair-ml approach may not result in the fair distribution of outcomes that motivated its use and can even harm those the intervention was designed to protect. Our arguments illustrate that fairness interventions are only meaningful if it carefully addresses the moral and empirical assumptions which motivate a particular distribution of outcomes.

To the best of our knowledge, we are the first to explicitly extend moral reasoning from fairness metrics to the fair-ml algorithms that optimize for them.
We believe this to be an important research direction that so far has been overlooked in philosophical work on algorithmic fairness. 

Our results show that fair-ml techniques are by no means the only part of the puzzle towards fairer machine learning systems. In an unequal \textit{status quo}, we often cannot meaningfully address the root causes of inequality at the level of a single decision-making point. Moreover, in many cases, posing fairness as an optimization problem is not the most effective way towards fairer outcomes. In particular, several scholars have highlighted how efforts such as more thoughtful measurement \citep{Jacobs2021}, expanding our view beyond the algorithmic frame \citep{corbett2017algorithmic, Selbst2018, green2021escaping}, and meaningful participatory design \citep{sloane2020participation} can provide critical paths towards fairer machine learning systems. That being said, when fair-ml approaches \textit{are} adopted, it is crucial to carefully consider the moral implications of doing so.

Importantly, decisions conceptualized as `technical' are often left at the discretion of the machine learning practitioner~\citep{wong2020democratizing}. Our analysis has shown that the `technical' choice of a fair-ml algorithm has moral implications. This results in a problem of legitimacy: do machine learning practitioners have the moral authority to decide what is a just distribution of burdens and benefits? Following \citep{daniels2007just}, we suggest that efforts towards distributive justice must therefore be complemented with forms of procedural justice, including transparency regarding the (moral) reasoning behind design choices. The present paper provides an important step in this direction, but much work remains to be done.

An important limitation of our work is the use of highly stylized examples in which we have only considered the direct short-term impact of the machine learning model in isolation. Reality is much more complex, including downstream or cumulative effects on individuals and communities \citep{Selbst2018}. In the long term, we envision a set of guidelines that support practitioners in considering and communicating the moral implications of particular strategies in various contexts, as well as new fair-ml algorithms tailored towards more nuanced empirical and moral assumptions.
 

\bibliographystyle{ACM-Reference-Format}
\bibliography{bibliography}


\end{document}